# Detection and Visualization of Endoleaks in CT Data for Monitoring of Thoracic and Abdominal Aortic Aneurysm Stents

J. Lu[ab], J. Egger[ac], A. Wimmer[ab], S. Großkopf*[a], B. Freisleben[c]

[a]Computed Tomography, Siemens Medical Solutions,
Siemensstraße 1, D-91301 Forchheim, Germany;
[b]Institute of Pattern Recognition, Friedrich-Alexander University of Erlangen-Nuremberg,
Martensstraße 3, D-91058 Erlangen, Germany;
[c]Dept. of Mathematics and Computer Science, Philipps-University of Marburg,
Hans-Meerwein-Straße, D-35032 Marburg, Germany

**ABSTRACT**

In this paper we present an efficient algorithm for the segmentation of the inner and outer boundary of thoratic and abdominal aortic aneurysms (TAA & AAA) in computed tomography angiography (CTA) acquisitions. The aneurysm segmentation includes two steps: first, the inner boundary is segmented based on a grey level model with two thresholds; then, an adapted active contour model approach is applied to the more complicated outer boundary segmentation, with its initialization based on the available inner boundary segmentation. An opacity image, which aims at enhancing important features while reducing spurious structures, is calculated from the CTA images and employed to guide the deformation of the model. In addition, the active contour model is extended by a constraint force that prevents intersections of the inner and outer boundary and keeps the outer boundary at a distance, given by the thrombus thickness, to the inner boundary. Based upon the segmentation results, we can measure the aneurysm size at each centerline point on the centerline orthogonal multiplanar reformatting (MPR) plane. Furthermore, a 3D TAA or AAA model is reconstructed from the set of segmented contours, and the presence of endoleaks is detected and highlighted. The implemented method has been evaluated on nine clinical CTA data sets with variations in anatomy and location of the pathology and has shown promising results.

## 1. INTRODUCTION

Cardiovascular diseases are the number one cause of death in the US and most European countries. Among these diseases, abdominal aortic aneurysm (AAA) is the 13th leading cause of death in the western world.

Endovascular aneurysm repair (EVAR) is a modern treatment alternative for aortic aneurysms which yields results comparable to classical open surgery[1, 2]. During EVAR, physicians deploy an endovascular stent graft in the aneurysm to replace and strengthen the weak aorta wall. However, incomplete exclusion of an aneurysm due to stent shift or deformation results in endoleaks: the leakage of blood around the stent graft and within the aneurysm sac. Endoleaks may cause continued pressurization of the aneurysm sac, which again increases the risk of aorta rupture.

To prevent further risks after EVAR and to determine the response of the aneurysm to the implanted stent graft, patients are required to take regular computed tomography angiography (CTA) scans of the abdominal region during the follow-up period. CTA can be very useful for obtaining exact knowledge of the position, shape, and size of the aneurysm and stent graft, and the occurrence of endoleaks. In practice, most of the aneurysm volume assessment is obtained by manual delineation slice by slice, which is laborious and non-reproducible. This paper presents an efficient algorithm for segmenting the inner and outer boundary – namely the lumen and thrombus boundary – of aneurysms in CTA images (Fig 1). Based on this segmentation, the aneurysm size, which is an important indicator for the risk of a rupture[12], can be measured on each MPR plane that is set orthogonal to the lumen centerline. Furthermore, a 3D model of the aneurysm can be reconstructed and visualized, and endoleaks can be detected.

*stefan.grosskopf@siemens.com; phone +49 9191 18-0; fax +49 9191 18-9990; siemensmedical.com

Several algorithms have been proposed to support clinicians with the evaluation of CTA scans during stent planning[3, 4, 5] and follow-up[6]. De Bruijne et al.[6] have presented a method for segmenting tubular structures based on active appearance models (AAM) and have applied it to AAA segmentation. Spreeuwers et al.[7] have introduced an approach to segment the epi- and endocard with the help of two ACMs and a constraint to keep the contours at a certain distance.

The method presented in this paper is based on the approach introduced by De Bruijne et al.[6] and applies it to TAA and AAA. Based on the segmentation results, the algorithm measures the size of the aneurysm for each centerline-orthogonal MPR plane and detects endoleaks in the thrombus volume. Furthermore, it reconstructs a 3D aneurysm model from the set of segmented contours and highlights the presence of endoleaks in 2D slices to attract the physician's attention. Compared to AAMs, our method does not require intensive training of a point distribution and texture model. The results obtained with the proposed method demonstrate the possibility of achieving efficient and precise segmentation of AAA and TAA thrombus and lumen.

The paper is organized as follows. Section 2 presents the details of the proposed method. Section 3 discusses the results of our experiments. Section 4 concludes the paper.

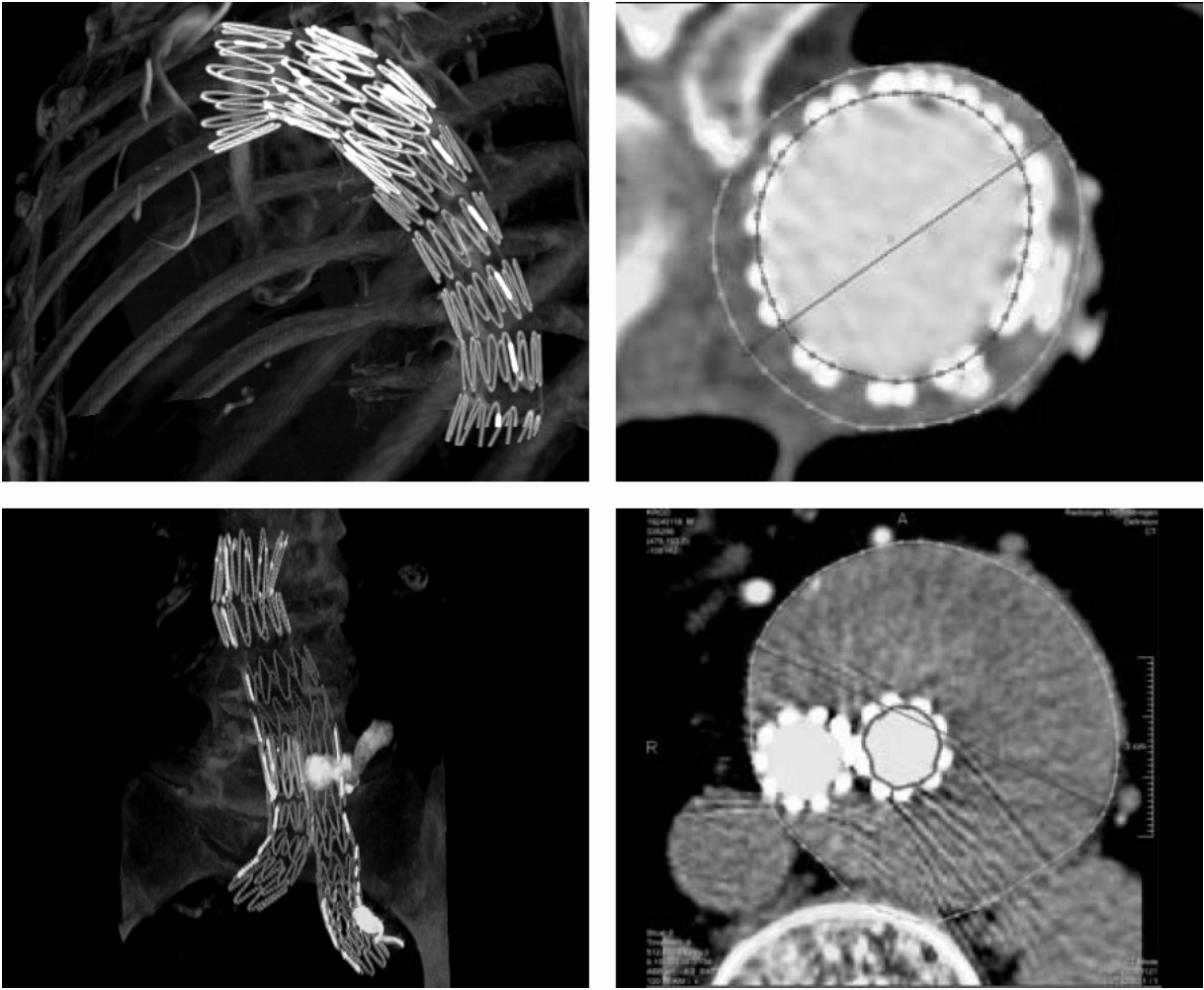

Fig. 1. Left: Postoperative CT-scan of a TAA stent graft (top) and an AAA stent graft (bottom). Right: Inner (Lumen) and outer (Thrombus) aneurysm boundary on a MPR plane of a TAA (top) and an AAA (bottom). The bright spots around inner contours are the metal parts of the stent grafts. Additionally, the maximum diameters of the outer contours are drawn.

## 2. METHODS

### 2.1 Segmentation of the Lumen Boundary

This paper extends our methods from previous work[8]. At first, the lumen centerline is determined by a vessel tracing method based on Dijkstra's shortest path algorithm[9, 10]. Then, the inner contours of the aneurysm are obtained by geometrical construction and 1D image processing. For this purpose, rays are sent out radially with equal angular spacing on centerline orthogonal planes (Fig. 2.), which are defined by the equidistant centerline points and the local centerline tangent vectors. The ray propagation stops as soon as the intensity value is below a threshold $\theta_1$. Additionally, this set of initial contours defined by the end points of the ray propagation is regularized using a simple threshold based 3D ACM approach.

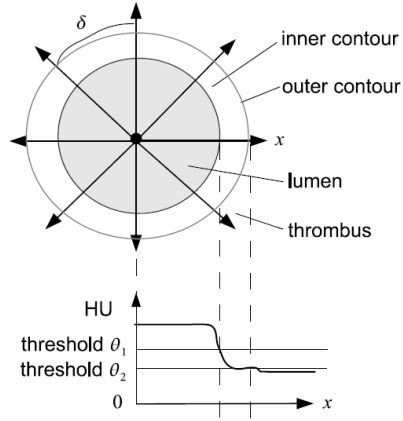

Fig. 2. Top: Equal angular spaced rays ($\delta$) propagate from the centerline point in a MPR slice. Bottom: Intensity profile from the center point along direction $x$; the threshold $\theta_1$ is defined as the lower threshold to detect the lumen boundary. Thresholding, even with adapted $\theta_2$, is usually not sufficient for detecting the weak thrombus boundary.

### 2.2 Segmentation of the Thrombus Boundary

The outer boundary of an aneurysm is segmented by an adapted 3D ACM approach. The ACM is initialized by dilating the inner contours to positions near the outer boundary. It is then deformed by internal and external forces to reveal the weak and sometimes incompletely imaged thrombus boundaries. The definition of the ACM's external energy for the outer contour segmentation is based on opacity images[11], which are calculated from the CTA image. An opacity image includes the information of both intensity value and gradient magnitude of a local point, aiming at enhancing important features while reducing spurious structures (Fig. 3).

The image force at a point $v \in R^3$ is calculated by convolving the opacity image $\alpha(x)$ with the derivative of a Gaussian kernel $G'_\sigma$ along a ray from the centerline point through $v$:

$$F_{image}(v) = G'_\sigma * \alpha(v) \qquad (1)$$

$\sigma$ is the standard deviation of the Gaussian kernel. It is reduced during the ACM deformation, leading to a multi-scale approach.

Additionally, a constraint force is employed, based on the local average distance between the inner and outer contour, to approximate a smooth boundary at positions where the opacity image has almost no information due to neighboring tissues with similar contrast:

$$F_{con}(v) = w_{con}\left(d_{min}(v) - \overline{d_N}(v)\right)n(v) \qquad (2)$$

Here, $w_{con}$ is a constant positive weighting parameter, $d_{min}(x)$ represents the shortest distance from point $x$ on the outer contour to the inner contour, $d_N(x)$ is the average distance between two corresponding points on outer and inner contour in a local neighborhood of $x$, and $n(x)$ is the inward unit normal at $x$.

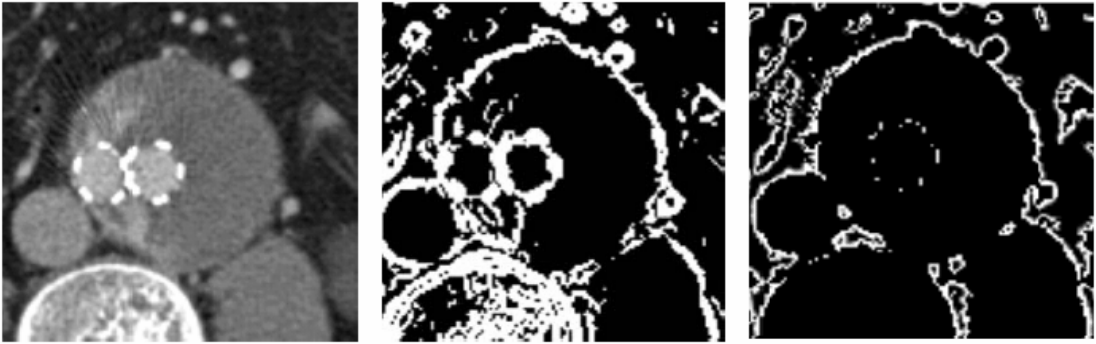

Fig. 3. Left: CTA slice containing AAA with implemented stent. Middle: Gradient image of the CTA slice with strong gradients in the lumen, stent metal markers and endoleak boundary. Right: Opacity image of the CTA slice where the strong misleading gradients are reduced or eliminated.

**2.3 Aneurysm Measurement**

As pointed out by Czermak et al.[12], the size of the aneurysm is an important indicator for the risk of a rupture. Therefore, two reproducible parameters are calculated from the outer boundary: the maximum aneurysm diameter $D_{max}$ and the maximum aneurysm cross-sectional area $A_{max}$. These two parameters are measured on each MPR plane perpendicular to the lumen centerline, which in general is more accurate than measuring on the original axial 2D slices due to the curved vessel.

**2.4 Endoleak Detection**

To detect endoleaks that are visible due to the contrast agent, the existence of a cluster of voxels for which the intensity values are above a threshold $\theta$ in the thrombus volume of CTA images is determined.

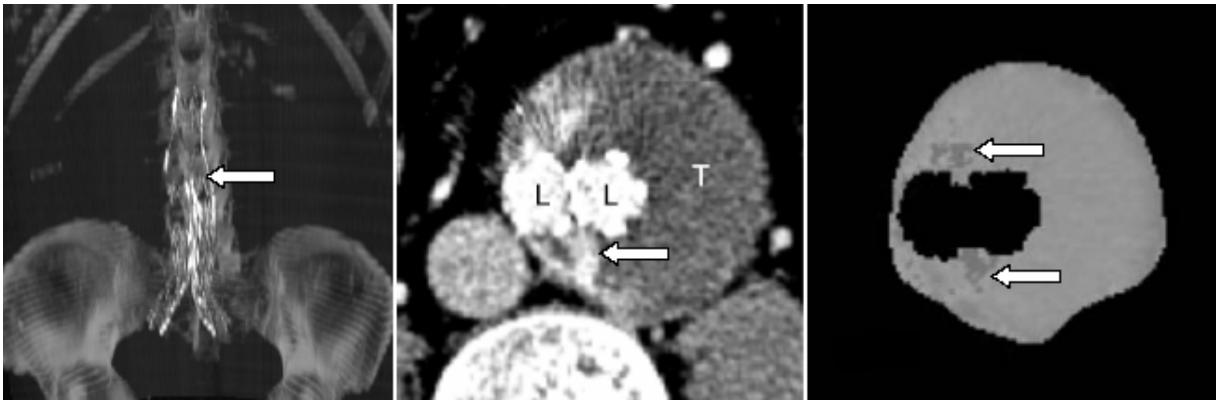

Fig. 4. Left: Bifurcated stent graft (white arrow). Middle: Slice with lumen (L), thrombus (T), stent graft (bright spots around lumen), endoleaks (white arrow) and neighboring structures. Right: Segmented thrombus volume with potential endoleaks around stent graft (white arrows).

A thrombus mask for analyzing HU intensities in the thrombus region is obtained by subtracting the lumen mask (rendered by voxelization of the triangulated inner contour) from the aneurysm volume (rendered by voxelization of the triangulated outer contour). Additionally, the metal markers of the stents and boundary voxels with high intensities due to the partial volume effect (PVE) have to be removed. Therefore, voxels whose intensity values are higher than a predefined metal threshold $\theta_s$ are marked in a binary stent mask. In order to eliminate remaining PVE voxels and get the desired volume, the binary stent mask is dilated and subtracted from the thrombus volume. Finally, intensity values that are greater than $\theta$ and still appear in the thrombus mask are candidates for occurring endoleaks (Fig. 4).

## 3. RESULTS

For our evaluation, the methods were implemented in MeVisLab[1]. Nine different clinical data sets were segmented (7 AAA scans and 2 TAA scans). Manual segmentations of the thrombus boundary were obtained by two trained observers and checked by a radiologist. The contours drawn manually on each slice were used to reconstruct a 3D surface and volume. The surface was used to generate an Euclidean distance map and the volume was used to create a binary mask. The surface and binary mask were adopted as the reference segmentation for the evaluation. Table 1 and Fig. 5 summarize the results obtained from the experiments for 9 scans; Fig. 6 and Fig. 7 show the segmentation results for a TAA and an AAA at different centerline positions.

Table 1. Summary of results: mean $\mu$, standard deviation $\sigma$, minimum and maximum values for 9 scans. $D_i$ is the segmentation distance error and DSC is the Dice Similarity Coefficient.

| Measure | $\mu \pm \sigma$ | [min, max] |
|---|---|---|
| Mean segmentation error ($\mu_{Di}$, mm) | 1.62 ± 1.03 | [0.06, 3.22] |
| Maximum distance ($\max_i D_i$, mm) | 8.04 ± 4.66 | [2.01, 16.3] |
| $D_i < 1$ mm (% vertices) | 56.5 ± 26.8 | [21.4, 98.9] |
| $D_i < 2$ mm (% vertices) | 74.6 ± 20.9 | [39.9, 100] |

| Case | 1 | 2 | 3 | 4 | 5 | 6 | 7 | 8 | 9 |
|---|---|---|---|---|---|---|---|---|---|
| DSC | 94.9 | 96.4 | 98.5 | 87.8 | 95.3 | 94.3 | 91.0 | 89.7 | 93.7 |

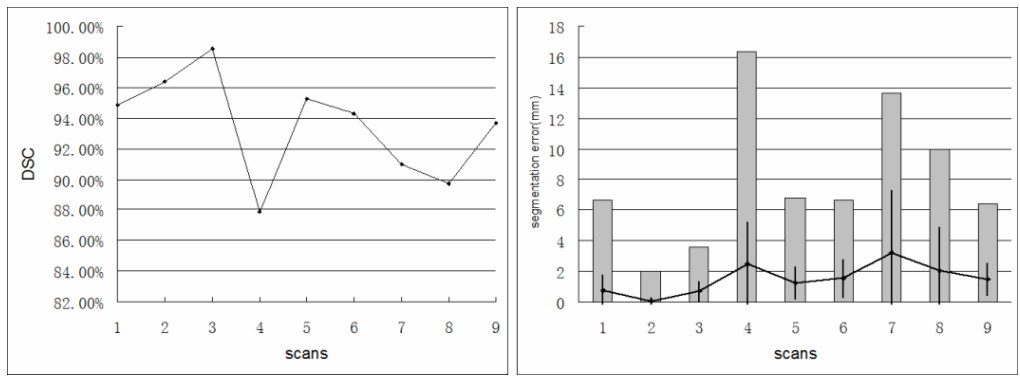

Fig. 5. Left: the Dice Similarity Coefficient (DSC) for 9 scans. Right: Segmentation error ($D_i$ in mm) for 9 scans. For $D_i$: maximum (bars), mean (dots connected by lines) and standard deviation (vertical lines inside).

---

[1] MeVisLab, Software for Medical Image Processing and Visualization, http://www.mevislab.de

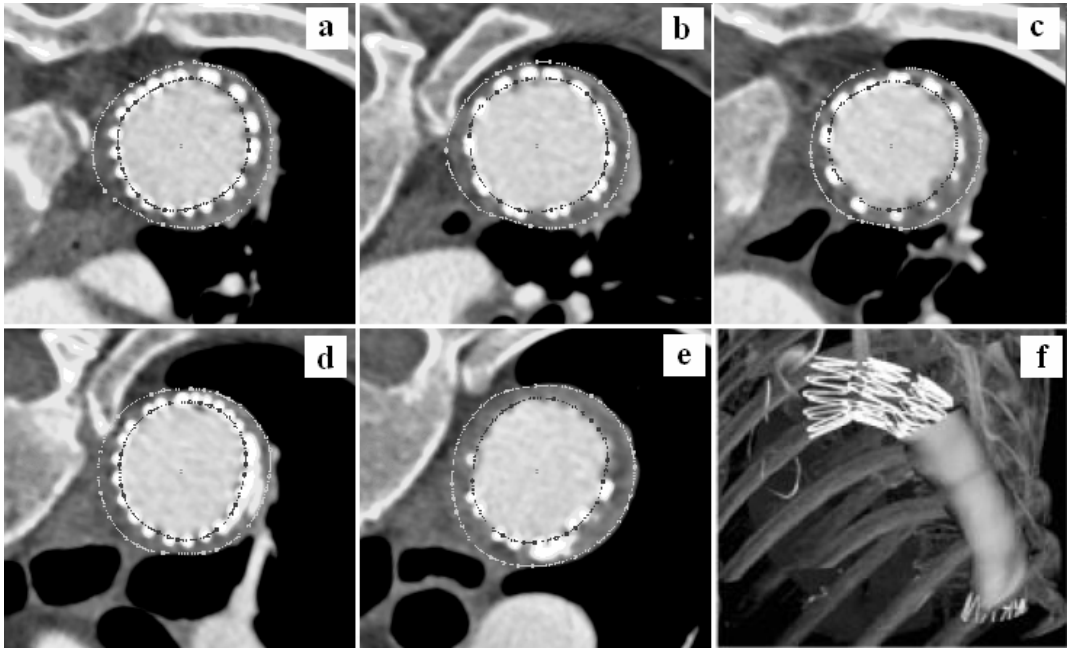

Fig. 6. Segmentation results of a TAA. Images (a) - (e) illustrate the segmented outer contour (Thrombus) and the inner contour (Lumen) on five different MPR planes. Image (f) visualizes the 3D aneurysm model with the outer boundary.

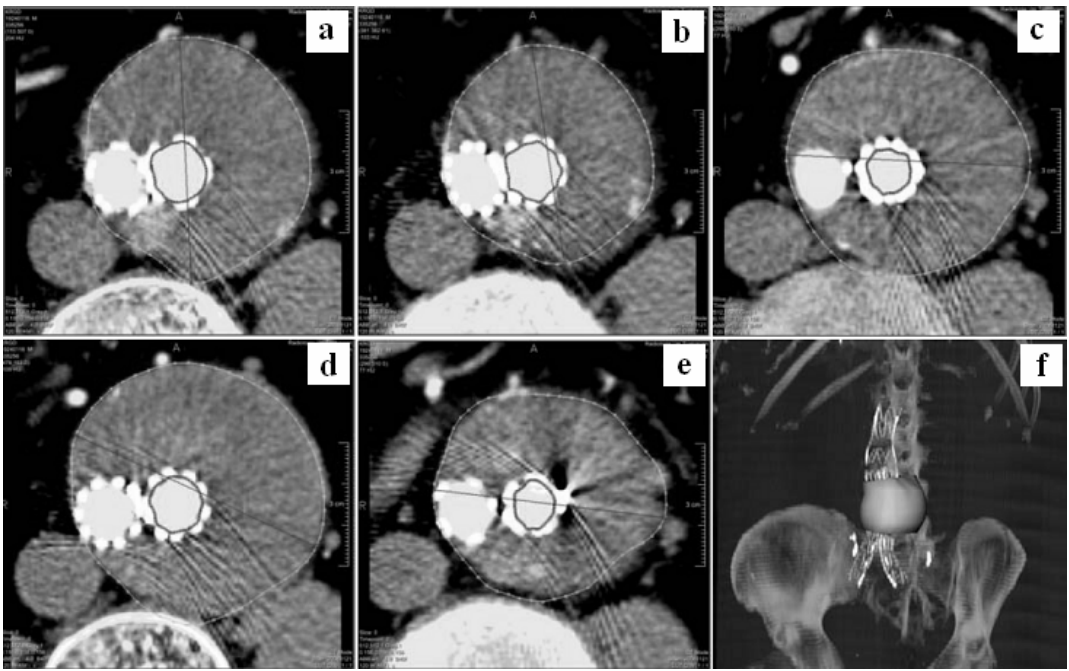

Fig. 7. Segmentation results of an AAA. Images (a) - (e) illustrate the segmented outer contour (Thrombus) and the inner contour (Lumen) on five different MPR planes. Image (f) visualizes the 3D aneurysm model with the outer boundary.

## 4. CONCLUSIONS

In this paper, an efficient algorithm for the segmentation of the inner and outer boundary of thoracic and abdominal aortic aneurysms in CTA acquisitions was presented, which saves analysis time and increases the reproducibility in treatment examination. Based on the segmentation, the aneurysm size was measured for each centerline-orthogonal MPR plane, and endoleaks were detected in the thrombus volume. Both indicate the risk of aneurysm rupture after an EVAR treatment. Furthermore, a 3D aneurysm model was reconstructed from the set of segmented contours, and the presence of endoleaks was highlighted in 2D slices to attract the physician's attention. Results obtained with the proposed method demonstrate the possibility of achieving efficient and precise segmentation of AAA and TAA thrombus and lumen. Thus, the method is an efficient alternative to manual segmentation and is useful for the measurement of AAA and TAA volume, allowing the assessment of aneurysm rupture risk in a more convenient manner.

## REFERENCES

[1] J. D. Blankensteijn, "Impact of the EVAR-1 and DREAM Trials", Endovascular Today March (2005).
[2] J. D. Blankensteijn, S. E. C. A. de Jong, M. Prinssen, A. C. van der Ham, J. Buth, S. M. M. van Sterkenburg, H. J. M. Verhagen, E. Buskens and D. E. Grobbee, "Two-Year Outcomes after Conventional or Endovascular Repair of Abdominal Aortic Aneurysms", N Engl J Med, 352:2398 (2005).
[3] J. Egger, S. Großkopf and B. Freisleben, "Simulation of bifurcated stent grafts to treat abdominal aortic aneurysms (AAA)", in Proceedings of SPIE Medical Imaging Conference, Vol. 6509, San Diego, USA (2007).
[4] J. Egger, S. Großkopf and B. Freisleben, "Preoperative Simulation of Tube- and Y-Stents for Endovascular Treatment of Stenosis and Aneurysms" (in German), Proceedings of Bildverarbeitung für die Medizin (BVM), Munich, Germany, pp. 182-186, Springer-Verlag, March (2007).
[5] J. Egger, Z. Mostarkic, S. Großkopf and B. Freisleben, "Preoperative Measurement of Aneurysms and Stenosis and Stent-Simulation for Endovascular Treatment", IEEE International Symposium on Biomedical Imaging: From Nano to Macro, Washington (D.C.), USA, pp. 392-395, IEEE Press, April (2007).
[6] M. de Bruijne, B. van Ginneken, W. J. Niessen and M. A. Viergever, "Adapting active shape models for 3D segmentation of tubular structures in medical images", Information Processing in Medical Imaging, Ambleside, UK, Vol. 2732, pp. 136-147 (2003).
[7] L. J. Spreeuwers and M. Breeuwer, "Myocardial boundary extraction using coupled active contours", Computers in Cardiology, Thessaloniki Chaldiki, Greece, pp. 745- 748 (2003).
[8] J. Lu, J. Egger, A. Wimmer, S. Großkopf and B. Freisleben, "Segmentation and Visualization of Lumen and Thrombus of Thoracic Aortic Aneurysm" (in German), Proceedings of 6. Jahrestagung der Deutschen Gesellschaft für Computer- und Roboterassistierte Chirurgie (CURAC), Karlsruhe, Germany, pp. 251-254, October (2007).
[9] J. Egger, Z. Mostarkic, S. Großkopf and B. Freisleben, "A Fast Vessel Centerline Extraction Algorithm for Catheter Simulation", 20th IEEE International Symposium on Computer-Based Medical Systems, Maribor, Slovenia, pp. (177-182), IEEE Press, June (2007).
[10] T. Boskamp, D. Rinck, F. Link, B. Kuemmerlen, G. Stamm and P. Mildenberger, "A New Vessel Analysis Tool for Morphometric Quantification and Visualization of Vessels in CT and MR Imaging Data Sets", Radiographics, 24:287-297 (2004).
[11] M. Levoy, "Display of surfaces from volume data", IEEE Computer Graphics & Applications, 8(5):29-37 (1988).
[12] B. Czermak, G. Fraedrich, M. Schocke, I. Seingruber, P. Waldenberger, R. Perkmann, M. Rieger, W. Jaschke, "Serial CT Volume Measurements After Endovascular Aortic Aneurysm Repair", Journal of Endovascular Therapy, Vol. 8, 4:380–389 (2001).